\documentclass{article}
\usepackage{ijcai25}

\usepackage{times}
\usepackage{soul}
\usepackage{url}
\usepackage[hidelinks]{hyperref}
\usepackage[utf8]{inputenc}
\usepackage[small]{caption}
\usepackage{graphicx}
\usepackage{amsmath}
\usepackage{amsthm}
\usepackage{booktabs}
\usepackage{algorithm}
\usepackage{algorithmic}
\usepackage[switch]{lineno}
\usepackage{multirow}
\usepackage{array}
\usepackage{xcolor}

\usepackage{amssymb}

\DeclareMathOperator*{\argmin}{arg\,min}
\DeclareMathOperator*{\argmax}{arg\,max}

\title{D\textsubscript{3}: Diversity, Difficulty, and Dependability-Aware Data Selection\\ for Sample-Efficient LLM Instruction Tuning}

\author{
Jia Zhang$^{1,2,3}$\thanks{Work done during the authors' internship at Alibaba Group.\\Preprint.}\and
Chen-Xi Zhang$^{1,2,3}$\footnotemark[1]\and
Yao Liu$^3$\and
Yi-Xuan Jin$^3$\and
Xiao-Wen Yang$^{1,2}$\and\\
Bo Zheng$^3$\and
Yi Liu$^3$\And
Lan-Zhe Guo$^{1,4}$\\
\affiliations
$^1$National Key Laboratory for Novel Software Technology, Nanjing University\\
$^2$School of Artificial Intelligence, Nanjing University\\
$^3$Algorithm Tech, Taobao \& Tmall Group of Alibaba\\
$^4$School of Intelligence Science and Technology, Nanjing University\\
}
\begin{document}
\maketitle

\begin{abstract}
Recent advancements in instruction tuning for large language models (LLMs) suggest that a small, high-quality dataset can significantly equip LLMs with instruction-following capabilities, outperforming large datasets often burdened by quality and redundancy issues.
However, the challenge lies in automatically identifying valuable subsets from large datasets to boost both the effectiveness and efficiency of instruction tuning. 
In this paper, we first establish data selection criteria based on three distinct aspects of data value: diversity, difficulty, and dependability, and then propose the D\textsubscript{3} method comprising two key steps of scoring and selection. Specifically, in the scoring step, we define the diversity function to measure sample distinctiveness and introduce the uncertainty-based prediction difficulty to evaluate sample difficulty by mitigating the interference of context-oriented generation diversity. Additionally, we integrate an external LLM for dependability assessment. 
In the selection step, we formulate the D\textsubscript{3} 
weighted coreset objective, which jointly optimizes three aspects of data value to solve for the most valuable subset. The two steps of D\textsubscript{3} can iterate multiple rounds, incorporating feedback to refine the selection focus adaptively. 
Experiments on both public datasets and the real-world Taobao Live application demonstrate the effectiveness of D\textsubscript{3} in endowing LLMs with competitive or even superior instruction-following capabilities using less than 10\% of the entire dataset.
\end{abstract}

\section{Introduction}
With the success of large language models (LLMs)~\cite{brown2020language,touvron2023llamaa},  the application of data-driven AI has been significantly advanced across various industries and scenarios. In the LLM paradigm, large-scale pre-training on massive corpora equips models with knowledge and text generation capabilities based on next-token prediction~\cite{naveed2024comprehensive}. 
Moreover, several post-training methods and techniques~\cite{bai2022training,wang2024comprehensive,rafailov2024direct} could further enhance the model's alignment and performance on downstream tasks. In particular, instruction tuning~\cite{ouyang2022training,Rohan023alpaca}, which fine-tunes models with labeled pairs of instruction and response, plays a crucial role in enabling LLMs' instruction-following capabilities across various tasks.

Early studies on instruction tuning concentrated on creating extensive and cross-task datasets to enhance the ability of LLMs to follow instructions at a significant cost~\cite{wang2022supernaturalinstructions,xu2023wizardlm,wang2023selfinstruct}. However, recent work~\cite{zhou2023lima} has shown that a small, carefully curated set of high-quality data can effectively equip LLMs with instruction-following capabilities instead of relying on large datasets. This finding has prompted initial efforts in automated data selection~\cite{xia2024less,cao2024instruction,du2023modsa,li2024superfilteringa,li2024quantitya}, where researchers attempt to automatically identify the valuable subset of instruction data for fine-tuning. Despite their potential, these methods generally face three limitations:
1) Many approaches rely on a single metric for data selection, neglecting  comprehensive assessments of instruction data.
2) Due to the inherent difference between inter- and intra-sample metrics like diversity and quality, existing methods often heuristically consider each in separate stages with manual thresholds to select data, hindering the dynamic balance of different aspects.
3) Lack of feedback to adaptively refine the selection focus.

To address the challenges above, we first establish data selection criteria based on three distinct aspects: diversity, difficulty, and dependability, corresponding to inter-sample, sample-model, and intra-sample perspectives of data value, respectively.
Building upon these criteria, we propose D\textsubscript{3}, an effective data selection method consisting of two key steps: scoring and selection. In the scoring step, specifically, to cope with the commonly observed context-oriented generation diversity (CoGD) in text generation, we introduce the uncertainty-based prediction difficulty (UPD), a novel score that quantifies the prediction difficulty of each sample by mitigating the interference caused by CoGD. For diversity and dependability, we define the diversity function based on sample embeddings and use a teacher model to assess the sufficiency and reliability of both the sample's instruction and its annotated response.
In the selection step, we formulate the D\textsubscript{3} weighted coreset object, which jointly optimizes the three aspects of the data value to solve for the most valuable data subset. Additionally, the scoring and selection steps can iterate through multiple rounds. This allows for adaptive refinement of the selection focus by incorporating sample-level feedback from the model during instruction tuning.

We conducted extensive experiments on three instruction tuning datasets to validate the D\textsubscript{3} method. Remarkably, with only 5\% to 10\% of the entire dataset, our approach achieves competitive or even superior performance compared to full-dataset fine-tuning. This highlights the feasibility of D\textsubscript{3} for sample-efficient instruction tuning, which enhances both efficacy and efficiency, thereby advancing related research. To summarize, our main contributions are as follows:

\begin{enumerate}
    \item[(1)] \textbf{Problem}: We establish data selection criteria based on three distinct aspects: diversity, difficulty, and dependability, corresponding to inter-sample, sample-model, and intra-sample perspectives of data value, respectively, which offers systematic guidance for the problem of data selection to enable sample-efficient instruction tuning.
    \item[(2)] \textbf{Method}: We propose the D\textsubscript{3} approach, consisting of two key steps. In scoring, we define the diversity function, introduce the uncertainty-based prediction difficulty and integrate a teacher model for dependability assessment. In selection, we formulate the D\textsubscript{3} weighted coreset object to jointly optimize three criteria and solve for the most valuable data subset. The two steps can iterate multiple rounds to adaptively refine the selection focus.
    \item[(3)] \textbf{Evaluation}: Extensive experiments validate the effectiveness of our proposal, demonstrating its potential for sample-efficient instruction tuning to enhance both efficacy and efficiency, thereby facilitating related research.
\end{enumerate}

\section{Related Work}
\subsection{Instruction Tuning for LLMs}
Instruction tuning~\cite{brown2020language,touvron2023llamaa} has emerged as a widely adopted post-training strategy to equip language models with instruction-following capabilities for various downstream tasks.
Early research~\cite{khashabi2020unifiedqa,ye2021crossfit} primarily focused on manually creating task-specific instruction tuning datasets. More recent approaches~\cite{honovich2022unnatural,wang2022supernaturalinstructions,xu2023wizardlm,wang2023selfinstruct,Rohan023alpaca} shift toward constructing instruction tuning datasets by leveraging powerful LLMs to generate data. These studies facilitate the automatic data creation  via distilling knowledge from the teacher model, and are also related to the growing field of LLM-based data synthesis~\cite{wang2024survey,long2024llmsdriven}. These studies have laid the foundation for instruction tuning, demonstrating that large task-specific or cross-task instruction tuning datasets can boost the instruction-following performance of language models.

\subsection{Sample-Efficient Tuning}
However, large instruction tuning datasets are often plagued by concerns over quality and redundancy. Some studies~\cite{zhou2023lima} have suggested that a small, manually curated set of high-quality data can endow the model with powerful instruction-following capabilities. To improve both the performance and efficiency of instruction tuning, recent research attempts to select valuable subsets for training from large datasets. These efforts generally focus on quality and diversity assessments.
\cite{du2023modsa} employs a reward model to assess sample quality, while~\cite{he2024shed} introduces the use of Shapely value to measure the marginal contribution of samples to overall performance. \cite{li2024quantitya,li2024superfilteringa} propose the IFD score to measure the model's difficulty in following instructions of each sample. \cite{kung2023activea} adopts an active learning framework, selecting samples based on prompt uncertainty to enhance cross-task performance. 
For diversity, k-means or k-center~\cite{sener2018active} techniques are typically employed to select data~\cite{ge2024clustering,du2023modsa}. 
However, many approaches rely on a single metric for data selection, neglecting a comprehensive assessment of instruction samples. Additionally, due to the inherent difference between inter-sample metrics like diversity and intra-sample metrics like quality, existing methods typically consider each aspect separately to select data with manual thresholds, lacking a unified way to jointly optimize multiple evaluation criteria. Besides, most methods lack feedback mechanisms to refine the selection focus. Moreover, the evaluation process in some methods is computationally expensive, such as using LLMs to generate responses for all training data, resulting in inefficiencies.

\begin{figure*}[t]
    \centering
    \vspace{-10pt}
    \includegraphics[width=0.95\textwidth]{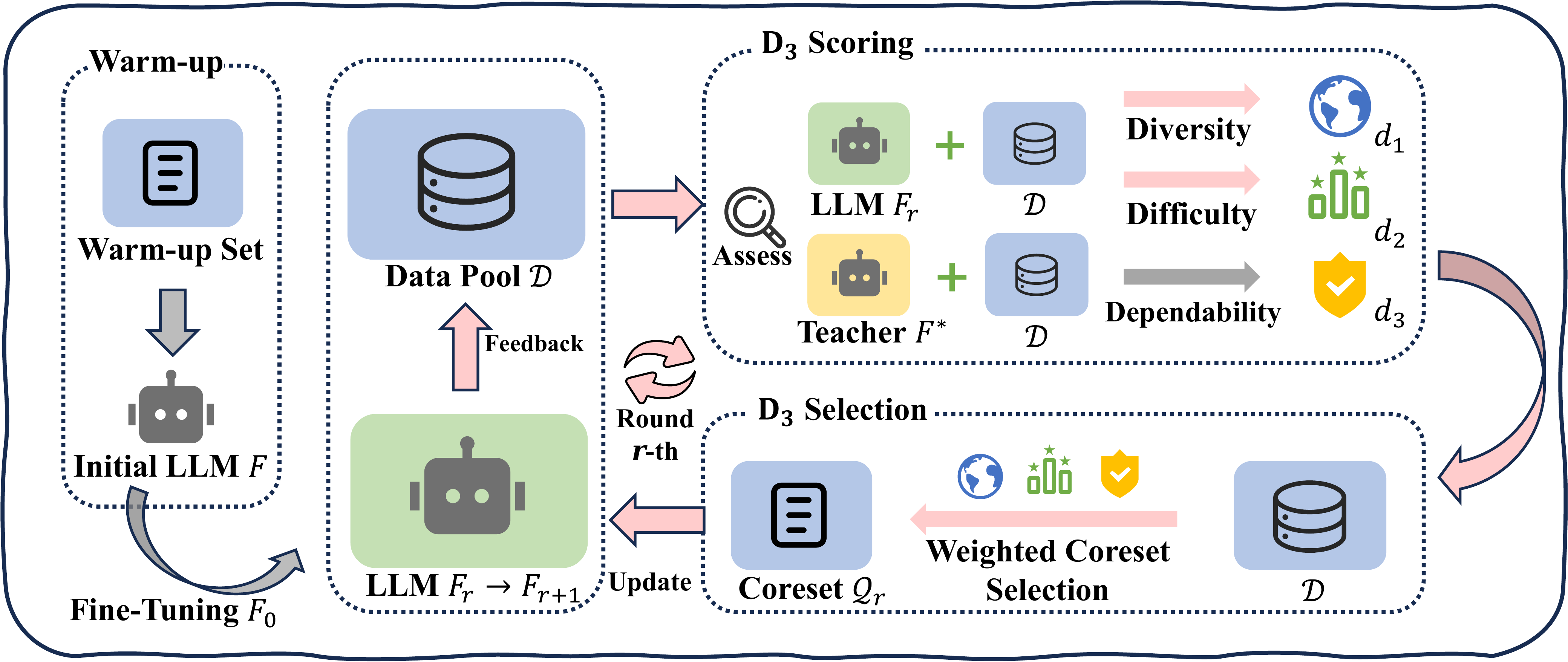}
    \caption{The overall framework of D\textsubscript{3} data selection method, including the warm-up and two key steps of scoring and selection. }
    \vspace{-10pt}
    \label{fig:framework}
\end{figure*}

\section{Methodology: \texorpdfstring{D\textsubscript{3}}{D3}}
\subsection{Problem Formalization}
In the task of data selection for the sample-efficient LLM instruction tuning, we are provided with an instruction tuning data pool $\mathcal{D}=\{z^{(i)}\}$ consisting of paired instruction and response samples $z^{(i)}=(x^{(i)}, y^{(i)})$, and a pre-trained LLM $F$ to be fine-tuned on $\mathcal{D}$. Our goal is to select a data coreset $\mathcal{Q}$ from the entire dataset $\mathcal{D}$ under a selection budget constraint of $|\mathcal{Q}|/|\mathcal{D}|= k$, such that the model $F_\mathcal{Q}$, fine-tuned from $F$ on $\mathcal{Q}$, achieves optimal performance on the test dataset $\mathcal{D}_t$.

\subsection{Overall Framework}
To tackle the challenge of identifying and selecting the most valuable coreset for instruction tuning, we first establish data selection criteria based on three distinct aspects: diversity, difficulty, and dependability, corresponding to inter-sample, sample-model, and intra-sample perspectives of data value, respectively. We then propose the novel D\textsubscript{3} method for selecting data based on these criteria. As illustrated in Figure~\ref{fig:framework}, our method comprises a warm-up step and two key steps of scoring and selection. We first utilize a tiny portion of data to obtain the initially acclimated model $F_0$. In the scoring step, we evaluate samples in $\mathcal{D}$ across three criteria. Specifically, we define the diversity function to measure sample distinctiveness and introduce the uncertainty-based prediction difficulty (UPD) to assess sample difficulty by mitigating the interference of context-oriented generation diversity. Additionally, we incorporate a teacher LLM $F^*$ to evaluate sample dependability. In the selection step, we formulate the D\textsubscript{3} weighted coreset objective to solve for the most valuable subset for LLM tuning within the selection budget. We integrate the scoring and selection steps into an iterative process that leverages the feedback from LLM to enable the adaptive refinement of the selection focus over multiple rounds.

\subsection{D\texorpdfstring{\textsubscript{3}}{3} Criteria and Scoring}
Unlike large-scale pre-training on vast data, we seek to equip LLMs with instruction-following capabilities using much fewer labeled samples.
To identify the most valuable data subset, we establish data selection criteria based on three distinct aspects of diversity, difficulty, and dependability. Diversity, from an inter-sample perspective, quantifies the distinctiveness of each sample against others. Difficulty, a sample-model metric, gauges the challenge for the language model in fitting and predicting a given sample. Dependability serves as an intra-sample criterion to assess whether the sample itself is sufficient and reliable enough for training. These criteria offer a comprehensive evaluation of instruction samples from three independent perspectives, providing the systematic guidance for data selection to identify the most valuable data coreset with maximal diversity, difficulty, and dependability.

\subsubsection{Sample Diversity}
The selection of the instruction tuning data coreset should account for its diversity characteristics and representativeness, ensuring that the selected samples accurately capture the overall distribution of the entire dataset $\mathcal{D}$. This helps mitigate the risk of the model over-fitting to specific task types. We define the diversity function $d_1$ of any sample $z^{(i)}$ as the distance to its closest point in the coreset $\mathcal{Q}$, which quantifies the distinctiveness of the sample:
\begin{equation}
\label{eq.sample diversity}
d_1(z^{(i)}, \mathcal{Q}) = \min_{z' \in \mathcal{Q}} \mathbf{d}(z^{(i)}, z'),
\end{equation}
where $\mathbf{d}(\cdot, \cdot)$ denotes the distance between two samples. Therefore, selecting a coreset with maximal diversity and representativeness with respect to $\mathcal{D}$ is equivalent to identifying $ \mathcal{Q} $ that minimizes the maximum diversity across all samples:
\begin{equation}
\argmin_{\mathcal{Q}\in\mathcal{D}}\max_{z \in \mathcal{D}} d_1(z, \mathcal{Q}) = \max_{z \in \mathcal{D}} \min_{z' \in \mathcal{Q}} \mathbf{d}(z, z').
\end{equation}

Minimizing $\max_{z \in \mathcal{D}} d_1(z, \mathcal{Q}) $ ensures the coreset $\mathcal{Q}$ closely reflects the entire dataset distribution by reducing the discrepancy between any sample from $\mathcal{D}$ and the coreset $\mathcal{Q}$.

\subsubsection{Sample Difficulty}

It is evident that the prediction difficulty for the model varies across different samples, meaning not all samples contribute equally to improving the model's instruction-following capability. Following the process of minimizing cross-entropy loss for each token during language model training, the sample-wise cross-entropy loss $\mathcal{L}^{(i)}$ directly quantifies the model's fit and prediction difficulty for any given sample $z^{(i)}$:

\begin{equation}
\label{eq.token loss}
\mathcal{L}_t^{(i)} = -\log(p(y_t^{(i)} \mid x^{(i)}, y_{<t}^{(i)})),
\end{equation}
\begin{equation}
\mathcal{L}^{(i)} = \frac{1}{|y^{(i)}|}\sum_t \mathcal{L}_t^{(i)},
\end{equation}
where $y_{<t}^{(i)}$ denotes the token sequence preceding the $t$-th token, $\mathcal{L}_t^{(i)}$ represents the loss of the $t$-th token given instruction $x^{(i)}$ and $y_{<t}^{(i)}$, and $|y^{(i)}|$ is the total number of tokens in $y^{(i)}$.
 
\begin{figure}[t]
    \centering
    \includegraphics[width=0.95\columnwidth]{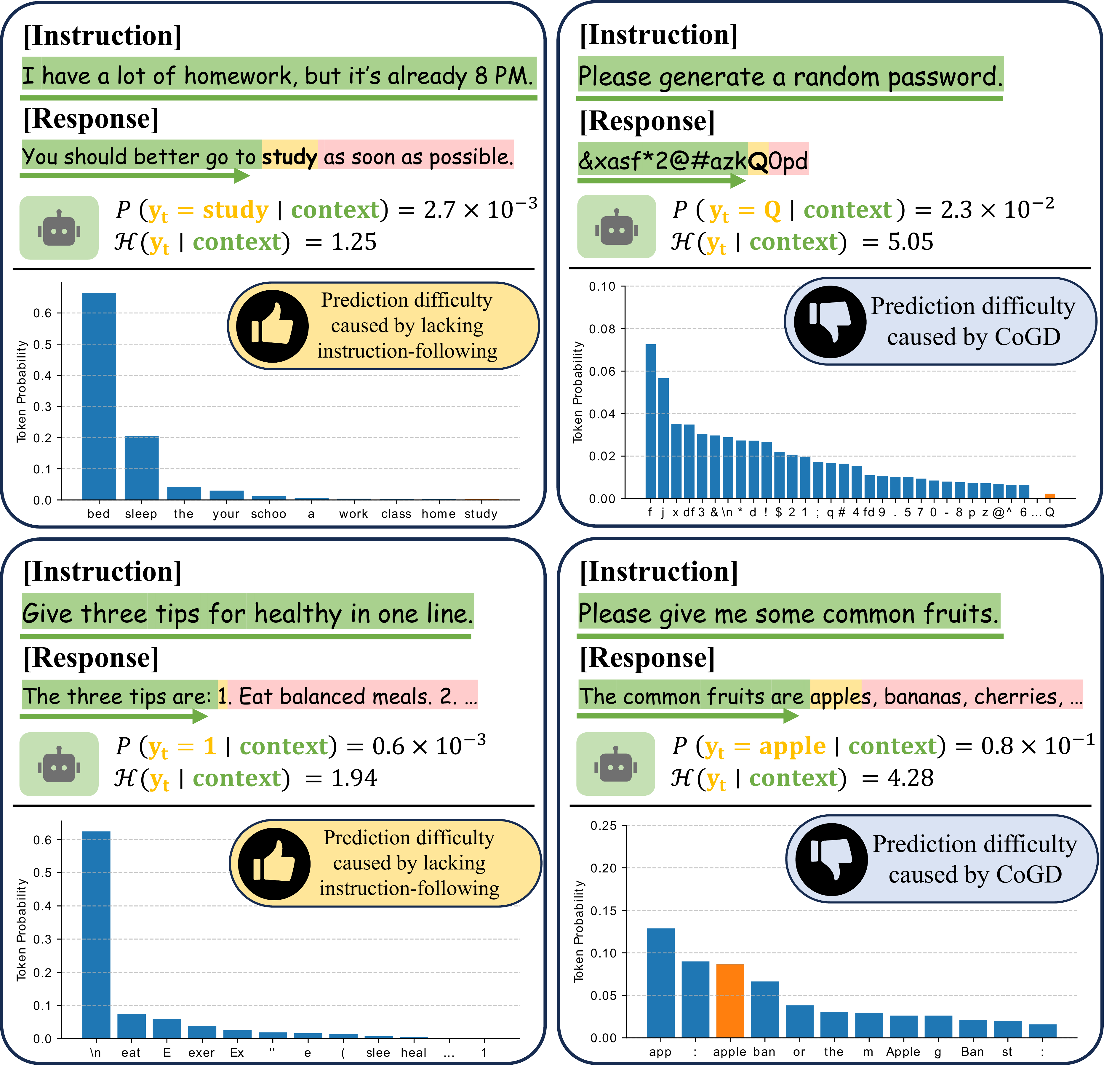}
    \caption{Token-wise analysis reveals two causes for tokens that are difficult to predict: context-oriented diversity and the model’s weak instruction-following, leading to confident but incorrect predictions. Instruction tuning should prioritize correcting poor instruction-following predictions rather than fitting difficult tokens caused by CoGD, as this risks reducing generation diversity and overfitting.}
    \vspace{-10pt}
    \label{fig:demonstration}
\end{figure}

However, improving an LLM's capability to follow instructions does not precisely align with minimizing loss on a limited amount of labeled instruction tuning data. As illustrated in Figure~\ref{fig:demonstration}, due to the dominant next-token prediction training and generation mechanism in current language models, inherent diversity in next-token generation, caused by the preceding context, commonly occurs in text generation tasks. We term this phenomenon context-oriented generation diversity (CoGD). A more intuitive understanding is that, for open-ended text generation, a single instruction can correspond to many valid responses. During large-scale pre-training, the vast data provides a broad coverage of this generation diversity, allowing the model to learn the distribution of this diversity from empirical data. In contrast, during instruction tuning, where the amount of training data is considerably reduced, overlooking CoGD in data selection risks over-fitting. Specifically, selecting samples with high loss caused by CoGD may lead the LLM to over-fit particular samples or tokens from distributions with greater generation diversity, thereby undermining its instruction-following capability.

To mitigate the interference of CoGD on the loss, which reflects sample difficulty, we introduce the uncertainty-based prediction difficulty (UPD), a metric that measures the model's prediction difficulty for samples by accounting for generation diversity. Since it is impractical to consider all potential responses to a given instruction at the sentence level, we quantify this by calculating the Shannon entropy~\cite{shannon1948mathematical} of token-wise predictions output from the LLM. This uncertainty reflects the generation diversity based on the current context and the pre-trained knowledge in LLM:
\begin{equation}
\label{eq.token h}
\mathcal{H}_t^{(i)}=-\sum_j p(w_j\mid x^{(i)},y_{<t}^{(i)})\log(p(w_j\mid x^{(i)},y_{<t}^{(i)})),
\end{equation}
where $\mathcal{H}_t^{(i)}$ is the entropy of the $t$-th token prediction given $x^{(i)}$ and $y_{<t}^{(i)}$, and $w_j$ denotes each token in the vocabulary.

Despite the high loss, tokens with higher uncertainty indicate greater context-oriented generation diversity. In data selection, we reduce the attention given to such tokens to mitigate the risk of over-fitting during instruction tuning. The UPD score for tokens and samples is computed as follows:

\begin{equation}
\label{eq.token upd}
\mathrm{UPD}^{(i)}_t=\sigma\left(\mathcal{L}_t^{(i)}\right)\cdot \max\left(1-\frac{\mathcal{H}^{(i)}_t}{[\log(v_{size})]^\beta}, 0\right),
\end{equation}
\begin{equation}
\label{eq.upd}
d_2(z^{(i)})=\text{UPD}(z^{(i)}) = \frac{1}{|y^{(i)}|} \sum_t \text{UPD}_t^{(i)},
\end{equation}
where $\sigma(u) = 2\left(\frac{1}{1 + e^{-{u}/{\alpha}}} - \frac{1}{2}\right) \in [0, 1]$ represents the sigmoid transformation mapping the loss score to $[0, 1]$, and $v_{\text{size}}$ is the size of the token vocabulary. Samples with a smaller UPD score typically exhibit either lower loss, indicating easier prediction, or higher uncertainty, reflecting greater generation diversity. These samples should be assigned lower weights to be considered and selected in the coreset.

\begin{algorithm}[t!]
\small
\caption{The D\textsubscript{3} Data Selection and Fine-tuning.}
\label{alg:d3}
\begin{algorithmic}[1]
    \STATE \textbf{Input:} Instruction tuning data pool $\mathcal{D}$, Pre-trained LLM $F$, Teacher LLM $F^*$, Selection budget $k$, Warm-up budget $k_0$, Total rounds $R$, Fine-tuning Algorithm $\mathcal{A}_{sft}$
    \STATE \textbf{Procedure:}
    \STATE Warm-up set $\mathcal{Q}_0\leftarrow \mathrm{sample}(\mathcal{D}, k_0)$ \COMMENT{Warm-up step}
    \STATE Warm-up LLM $F_0 \leftarrow \mathcal{A}_{sft}(F, \mathcal{Q}_0)$
    \FOR [D\textsubscript{3} scoring step]{round $r$ in $\{1\cdots R\}$} 
        \FOR {$z^{(i)} \in \mathcal{D}$}
            \STATE Calc. the sample embedding $\psi_{F_{r-1}}(z^{(i)})$ for the \\diversity function $d_1(z^{(i)}, \cdot)$ in Eq.(\ref{eq.sample diversity}).
            \STATE Calc. the difficulty score $d_2(z^{(i)})$ by Eq.(\ref{eq.token upd}), (\ref{eq.upd}).
            \STATE Calc. the dependability score $d_3(z^{(i)})$ by Eq.(\ref{eq.dependability}).
        \ENDFOR
        \STATE Initialize $\mathcal{Q}_r\leftarrow\varnothing$\COMMENT{D\textsubscript{3} selection step}
        \WHILE {$|\mathcal{Q}_r|/|\mathcal{D}| < {k}/{R}$}
            \STATE $\mathcal{Q}=\mathcal{Q}_1\cup\cdots\cup\mathcal{Q}_{r-1}\cup\mathcal{Q}_r$
            \STATE $z^*\leftarrow\argmax_z\{d_1(z, \mathcal{Q})\cdot d_2(z)\cdot d_3(z)\}$
            \STATE $\mathcal{Q}_r\leftarrow\mathcal{Q}_r\cup\{z^*\}$
        \ENDWHILE
        \STATE $F_r\leftarrow \mathcal{A}_{sft}(F_{r-1}, \mathcal{Q}_r)$ \COMMENT{Instruction tuning}
    \ENDFOR
\end{algorithmic}
\end{algorithm}

\begin{table*}[ht!]
\centering
\caption{Performance comparisons of various data selection strategies across different datasets. 
The numbers in parentheses denote the sample sizes of test sets.
Higher winning scores and leaderboard metrics indicate better performance, with the best results highlighted in \textbf{bold}.}
\small
\setlength{\tabcolsep}{5pt}
\vspace{-5pt}
\label{tab:exp1}
\begin{tabular}{@{}clcccccccc@{}}
\toprule
\midrule
\multicolumn{1}{l}{} &
  \multicolumn{1}{l|}{} &
  \multicolumn{6}{c|}{\textbf{Winning Score} (\textit{budget} $k$ vs. \textbf{Full})} &
  \multicolumn{2}{c}{\textbf{Leaderboard}} \\ \midrule
\multicolumn{1}{c|}{\begin{tabular}[c]{@{}c@{}}\textbf{Dataset}\\ (\textit{budget} $k$)\end{tabular}} &
  \multicolumn{1}{l|}{\begin{tabular}[c]{@{}l@{}}\textbf{Selection}\\ \textbf{Strategy}\end{tabular}} &
  \begin{tabular}[c]{@{}c@{}}WizardLM\\ (218)\end{tabular} &
  \begin{tabular}[c]{@{}c@{}}Sinstruct\\ (252)\end{tabular} &
  \begin{tabular}[c]{@{}c@{}}Vicuna\\ (80)\end{tabular} &
  \begin{tabular}[c]{@{}c@{}}Koala\\ (180)\end{tabular} &
  \multicolumn{1}{c|}{\begin{tabular}[c]{@{}c@{}}LIMA\\ (300)\end{tabular}} &
  \multicolumn{1}{c|}{\begin{tabular}[c]{@{}c@{}}Average\\ (1030)\end{tabular}} &
  \multicolumn{2}{c}{\begin{tabular}[c]{@{}c@{}}AlpacaEval\\ (\textit{budget} $k$ vs. \textbf{Full})\end{tabular}} \\ \midrule \midrule
\multicolumn{1}{c|}{\multirow{4}{*}{\begin{tabular}[c]{@{}c@{}}Alpaca\\ (\textit{5\%})\end{tabular}}} &
  \multicolumn{1}{l|}{\textbf{PPL}} &
  0.20 &
  \multicolumn{1}{>{\centering\arraybackslash}p{1.2cm}}{0.30} &
  \multicolumn{1}{>{\centering\arraybackslash}p{1.2cm}}{0.08} &
  \multicolumn{1}{>{\centering\arraybackslash}p{1.2cm}}{0.16} &
  \multicolumn{1}{>{\centering\arraybackslash}p{1.0cm}|}{0.15} &
  \multicolumn{1}{>{\centering\arraybackslash}p{1.5cm}|}{0.19} &
  \multicolumn{1}{>{\centering\arraybackslash}p{1.0cm}|}{\hspace{4pt}4.34} &
  \multirow{4}{*}{28.00} \\
\multicolumn{1}{c|}{} &
  \multicolumn{1}{l|}{\textbf{RAND}} &
  0.99 &
  0.93 &
  1.09 &
  0.98 &
  \multicolumn{1}{c|}{0.95} &
  \multicolumn{1}{c|}{0.97} &
  \multicolumn{1}{c|}{28.29} &
   \\
\multicolumn{1}{c|}{} &
  \multicolumn{1}{l|}{\textbf{IFD}} &
  1.15 &
  1.01 &
  \textbf{1.38} &
  1.13 &
  \multicolumn{1}{c|}{1.28} &
  \multicolumn{1}{c|}{1.17} &
  \multicolumn{1}{c|}{36.50} &
   \\
\multicolumn{1}{c|}{} &
  \multicolumn{1}{l|}{\textbf{D\textsubscript{3}}} &
  \textbf{1.22} &
  \textbf{1.16} &
  1.21 &
  \textbf{1.33} &
  \multicolumn{1}{c|}{\textbf{1.37}} &
  \multicolumn{1}{c|}{\textbf{1.27}} &
  \multicolumn{1}{c|}{\textbf{42.01}} &
   \\ \midrule
\multicolumn{1}{c|}{\multirow{4}{*}{\begin{tabular}[c]{@{}c@{}}AlpacaGPT4\\ (\textit{5\%})\end{tabular}}} &
  \multicolumn{1}{l|}{\textbf{PPL}} &
  0.29 &
  0.33 &
  0.11 &
  0.36 &
  \multicolumn{1}{c|}{0.20} &
  \multicolumn{1}{c|}{0.27} &
  \multicolumn{1}{c|}{20.92} &
  \multirow{4}{*}{65.68} \\
\multicolumn{1}{c|}{} &
  \multicolumn{1}{l|}{\textbf{RAND}} &
  0.85 &
  0.86 &
  0.83 &
  0.92 &
  \multicolumn{1}{c|}{0.98} &
  \multicolumn{1}{c|}{0.90} &
  \multicolumn{1}{c|}{60.93} &
   \\
\multicolumn{1}{c|}{} &
  \multicolumn{1}{l|}{\textbf{IFD}} &
  1.10 &
  1.08 &
  1.00 &
  1.11 &
  \multicolumn{1}{c|}{1.15} &
  \multicolumn{1}{c|}{1.10} &
  \multicolumn{1}{c|}{64.53} &
   \\
\multicolumn{1}{c|}{} &
  \multicolumn{1}{l|}{\textbf{D\textsubscript{3}}} &
  \textbf{1.13} &
  \textbf{1.19} &
  \textbf{1.07} &
  \textbf{1.24} &
  \multicolumn{1}{c|}{\textbf{1.22}} &
  \multicolumn{1}{c|}{\textbf{1.19}} &
  \multicolumn{1}{c|}{\textbf{71.26}} &
   \\ \midrule
\multicolumn{1}{c|}{\multirow{4}{*}{\begin{tabular}[c]{@{}c@{}}WizardLM\\ (\textit{5\%})\end{tabular}}} &
  \multicolumn{1}{l|}{\textbf{PPL}} &
  0.40 &
  0.52 &
  0.60 &
  0.45 &
  \multicolumn{1}{c|}{0.48} &
  \multicolumn{1}{c|}{0.48} &
  \multicolumn{1}{c|}{28.49} &
  \multirow{4}{*}{58.42} \\
\multicolumn{1}{c|}{} &
  \multicolumn{1}{l|}{\textbf{RAND}} &
  0.92 &
  0.81 &
  1.08 &
  0.79 &
  \multicolumn{1}{c|}{0.94} &
  \multicolumn{1}{c|}{0.89} &
  \multicolumn{1}{c|}{54.48} &
   \\
\multicolumn{1}{c|}{} &
  \multicolumn{1}{l|}{\textbf{IFD}} &
  0.93 &
  0.93 &
  \textbf{1.28} &
  0.99 &
  \multicolumn{1}{c|}{1.11} &
  \multicolumn{1}{c|}{1.02} &
  \multicolumn{1}{c|}{54.94} &
   \\
\multicolumn{1}{c|}{} &
  \multicolumn{1}{l|}{\textbf{D\textsubscript{3}}} &
  \textbf{0.99} &
  \textbf{1.14} &
  1.21 &
  \textbf{1.07} &
  \multicolumn{1}{c|}{\textbf{1.17}} &
  \multicolumn{1}{c|}{\textbf{1.11}} &
  \multicolumn{1}{c|}{\textbf{63.09}} &
   \\ \midrule\bottomrule
\end{tabular}
\end{table*}

It is worth noting that some uncertainty-based machine learning methods, such as in active learning~\cite{shao2022log,shao2022active1}, often prioritize samples with higher uncertainty, as these samples are less familiar to the model and more informative, which contrasts sharply with the motivation behind UPD. Unlike small models trained from scratch, LLMs are typically pre-trained on vast corpora, enabling them to acquire extensive knowledge and text generation capabilities~\cite{zhao2024survey}. Instruction tuning, which involves relatively much fewer training samples, primarily serves to equip the models with instruction-following capabilities rather than imparting new, low-level knowledge. Therefore, the priorities during instruction tuning should be correcting the model's confidently incorrect prediction patterns rather than reducing the context-oriented generation diversity. From this perspective, we observe that samples with higher UPD scores are primarily due to the model's weak instruction-following rather than factors related to CoGD. This suggests that the model is poorly aligned with these instruction samples, and such samples should be prioritized for instruction tuning to enhance the model performance.

\subsubsection{Sample Dependability}

The difficulty score based on UPD helps identify samples with greater difficulty that less likely to be caused by CoGD. However, if a sample itself contains factual errors, extraneous symbols, or lacks fluency, it can undermine both the data selection and instruction tuning. To tackle this, we integrate an external teacher model to assess the dependability of samples through prompt-based evaluation. This helps filter out unreliable samples, reducing the risk of incorporating erroneous supervision during instruction tuning. By leveraging the teacher's instruction-following capability, we obtain a soft dependability score $d_3$ within $(0, 1)$ for each sample:
{
\small
\begin{equation}
\label{eq.dependability}
d_3(z^{(i)})=\frac{\exp(F^*(w_{pos}\mid T(z^{(i)})))}{\exp(F^*(w_{pos}\mid T(z^{(i)})))+\exp(F^*(w_{neg}\mid T(z^{(i)})))},
\end{equation}
}
where $ F^* $ represents the teacher model, $ w_{\text{pos}} $ and $ w_{\text{neg}} $ correspond to the positive and negative tokens, respectively. $ T(z^{(i)}) $ is the evaluation prompt generated based on sample $ z^{(i)} $, and $ F^*(w | T) $ refers to the logit of the next token $ w $ given the inputs $ T(z^{(i)}) $. Details are provided in the appendix.

\subsection{D\texorpdfstring{\textsubscript{3}}{3} Selection Objective}
To enhance the model’s instruction-following capability efficiently, we select coreset data under a selection budget $k$ that captures the diversity of the overall data distribution while ensuring the samples in the coreset exhibit great difficulty and dependability. We formalize this multi-criteria sample selection problem as the D\textsubscript{3} weighted coreset objective, aiming to solve for the coreset $\mathcal{Q}$ which jointly optimizes the three aspects of selection criteria under the budget constraint:

\begin{align}
\mathcal{Q}&=\argmin_{\mathcal{Q}} \max_{z\in \mathcal{D}} d_1(z, \mathcal{Q}) d_2(z) d_3(z)\quad s.t. \frac{|\mathcal{Q}|}{|\mathcal{D}|}= k\nonumber\\
&=\argmin_\mathcal{Q}\max_{z\in \mathcal{D}}\min_{z'\in\mathcal{Q}}d_2(z)d_3(z)\mathbf{d}_{\psi_F}(z,z'),
\end{align}
where $\mathbf{d}_{\psi_F}$ represents the cosine distance between the embeddings of samples $\psi_F(\cdot)$ produced by the LLM $F$. We tackle this NP-hard combinatorial optimization problem with the greedy iterative approach. Specifically, we initialize $\mathcal{Q}$ as an empty set and randomly select a sample to add to $\mathcal{Q}$. In each iteration, we compute the weighted nearest distance between the remaining samples and $\mathcal{Q}$, selecting the sample with the largest distance to add to $\mathcal{Q}$. We repeat this process until the desired number of samples is obtained.

We conduct multiple rounds of scoring and selection steps. In each round $r$, we select the coreset $\mathcal{Q}_r$ based on the scoring results and then fine-tune the model to obtain $F_r$. Subsequently, leveraging the sample-level real-time feedback from $F_r$, we re-score and re-select the samples for subsequent fine-tuning, which enables the adaptive refinement of the selection focus. The process of D\textsubscript{3} is summarized in Algorithm~\ref{alg:d3}.

\section{Empirical Study}
\begin{table}[t]
\small
\centering
\caption{Pairwise comparisons on real-world Taobao Live dataset across five evaluation dimensions of interest: Satisfaction, Presentation Skill, Conversion, Spoken Fluency, and Overall Performance.}
\label{tab:exp.1.1}
\begin{tabular}{@{}|lccccc@{}}
\toprule
\multicolumn{1}{c|}{}        & \multicolumn{5}{c}{\textbf{Winning Score} (\textit{budget 10\%} vs.   \textbf{Full})}   \\ \midrule
\multicolumn{1}{l|}{\textbf{Strategy}} & Satis. & Pres. & Conv. & \multicolumn{1}{c|}{Spoke.} & Overall \\ \midrule
\multicolumn{1}{l|}{\textbf{PPL}}    & 0.87 & 0.84 & 0.81 & \multicolumn{1}{c|}{0.89} & 0.82 \\
\multicolumn{1}{l|}{\textbf{RAND}} & 0.90 & 0.91 & 0.92 & \multicolumn{1}{c|}{0.89} & 0.93 \\
\multicolumn{1}{l|}{\textbf{IFD}}    & 0.77 & 0.74 & 0.76 & \multicolumn{1}{c|}{0.82} & 0.71 \\
\midrule
\multicolumn{1}{l|}{\textbf{D\textsubscript{3}}}     & \textbf{1.07} & \textbf{1.07} & \textbf{1.04} & \multicolumn{1}{c|}{\textbf{1.06}} & \textbf{1.11} \\ \bottomrule
\end{tabular}
\end{table}

\subsection{Experimental Setup}

\begin{figure*}[t]
    \centering
    \vspace{-10pt}
    \includegraphics[width=0.9\textwidth]{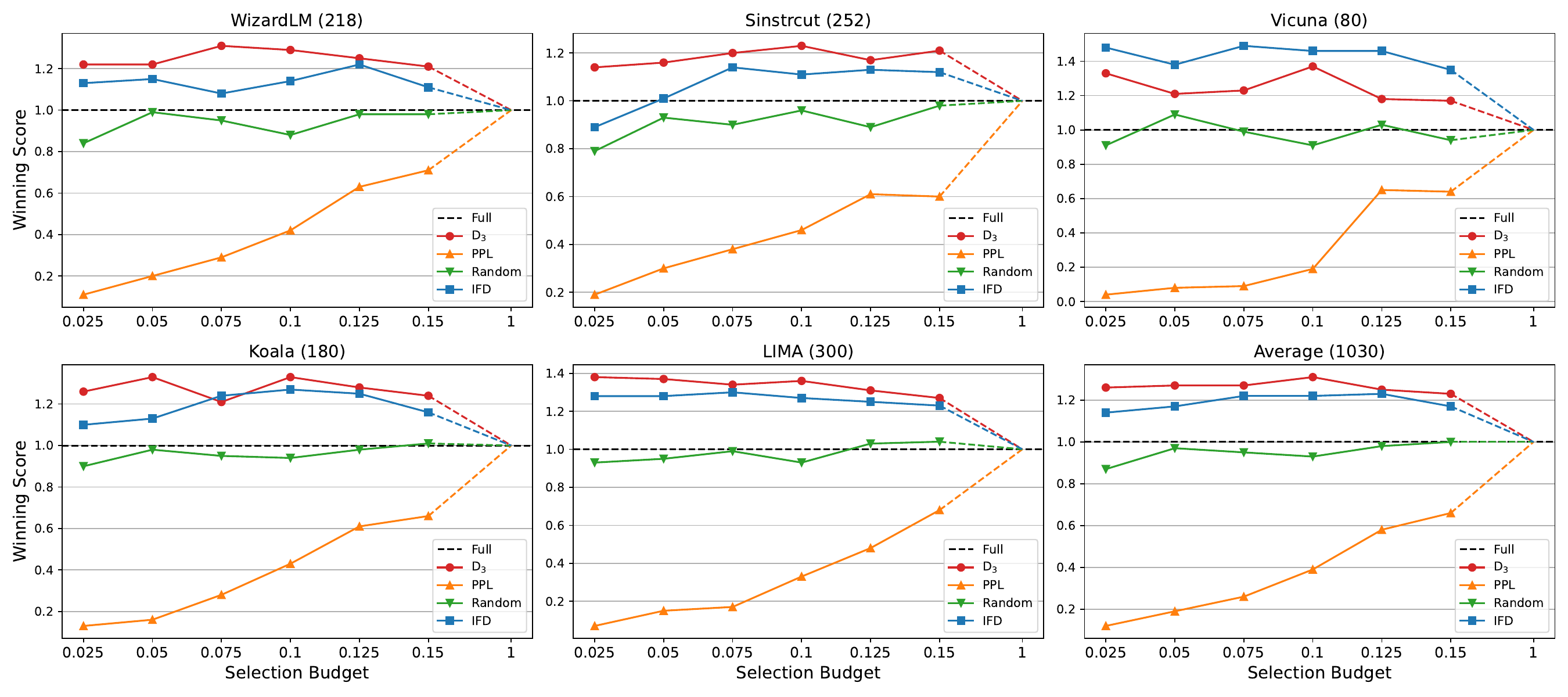}
    \vspace{-5pt}
    \caption{Performance variation across different selection budgets on Alpaca is presented. Detailed results can be found in the appendix.}
    \vspace{-10pt}
    \label{fig:exp2}
\end{figure*}

\textbf{Training Datasets.} We employ three commonly used instruction tuning datasets in the following experiments: 1) The Alpaca dataset~\cite{Rohan023alpaca}, generated through the self-instruct method~\cite{wang2023selfinstruct} and Davinci003 model, consists of 52,002 samples. Given that the data generation relies on Davinci003, the overall quality of the dataset is moderate, akin to datasets collected in real-world tasks that may include low-quality or erroneous samples. 2) The Alpaca-GPT4 dataset~\cite{peng2023instruction} is a refined version of the Alpaca dataset generated using GPT-4, resulting in significantly higher quality. 3) The WizardLM dataset~\cite{xu2023wizardlm}, created by the Evol-Instruct method with ChatGPT, consists of 70,000 high-quality instruction samples in total.

\noindent\textbf{Test Datasets.} We evaluate the effectiveness of different data selection strategies using five test datasets: WizardLM~\cite{xu2023wizardlm}, Self-instruct~\cite{wang2023selfinstruct}, Vicuna~\cite{vicuna2023}, Koala~\cite{vu2023koala}, and LIMA~\cite{zhou2023lima}. These datasets consist of approximately 1,000 instruction-based queries in total, curated across diverse tasks and domains. They offer a comprehensive evaluation of the models' capabilities to follow instructions in real-world tasks.

\noindent\textbf{Real-world Dataset.} We further evaluate our method in a real-world setting within Taobao Live, where the language model is fine-tuned using collected live-streaming data to generate high-quality sales-pitch scripts tailored to product information, thereby empowering streamers to improve their performance. We utilize 64,881 collected samples for training, with a random selection of 300 reserved for testing. A pairwise evaluation of generated scripts is conducted across five evaluation dimensions of interest in the task context.

\noindent\textbf{Candidate Methods.} 
We compare our proposed D\textsubscript{3} method with the following data selection strategies:
\begin{itemize}
    \item \textbf{Full}: No data selection strategy. The entire instruction tuning dataset is utilized to fine-tune the LLMs.
    \item \textbf{RAND}: A subset of training data is randomly sampled from the original instruction tuning dataset.
    \item \textbf{PPL}: The samples with the highest perplexity are selected, corresponding to those with the highest loss.
    \item \textbf{IFD}~\cite{li2024quantitya}: The IFD score, calculated as the ratio of loss to conditional loss, identifies samples with weaker instruction-following. The subset of samples with the highest IFD scores is then selected.
\end{itemize}
To ensure fairness and adherence to the settings and procedure of candidate methods, we use the pre-trained Llama2-7B as the initial model. All methods first undergo two training epochs on a warm-up set, which consists of 1\% of the data randomly sampled from the data pool. Subsequently, each strategy selects a data subset of equal size for instruction tuning based on a given selection budget. Samples selected are trained for three epochs to obtain the fine-tuned LLMs.

\subsection{Evaluation and Benchmark}
\textbf{Pairwise Evaluation.} Evaluating the instruction-following capabilities of LLMs on open-ended natural language generation tasks remains a significant challenge. Building on recent advancements in LLM evaluation, such as LLM-as-a-judge~\cite{zheng2023judging,li2025generation}, a more capable LLM is employed as a judge to assess response quality. We generate two responses for each instruction sample in the test set $\mathcal{D}_t$: one from the model fine-tuned on a small subset selected and the other from a model fine-tuned on the full dataset. The judge LLM assigns preference scores, and based on these scores, we classify the outcome as a win, tie, or loss for each sample. Each response pair is evaluated twice in different orders to mitigate positional bias. The final winning score is computed from outcomes of the entire test set. Pairwise evaluations are performed across all five test datasets.
\begin{equation}
    \mathrm{winning\_score}(\mathcal{D}_t)=\frac{\mathrm{num(wins)}-\mathrm{num(loses)}}{\mathrm{num}(\mathcal{D}_t)}+1
\end{equation}

\noindent\textbf{Benchmark.} We also evaluate models on the widely recognized AlpacaEval Leaderboard~\cite{alpaca_eval}, which provides automated assessments using the AlpacaFarm dataset~\cite{dubois2023alpacafarm}. This evaluation compares the model outputs with those of Davinci003 by ChatGPT. Additional details on the evaluation are provided in the appendix.

\subsection{Main Results}
We present the primary comparison results in Table~\ref{tab:exp1}. For all three instruction tuning datasets, we allocate a $k=5\%$ selection budget for sample-efficient instruction tuning. To ensure a fair comparison, all strategies select data to reach the budget in one go, except for the Full strategy, which uses all the data. Specifically, we set total round $ R = 1 $ in our D\textsubscript{3}. Our proposed D\textsubscript{3} method demonstrates superior results by effectively selecting a small, yet highly valuable subset of data, enhancing the model's instruction-following capabilities on both medium and high-quality instruction tuning datasets. It outperforms the Full strategy in both performance and efficiency. In contrast, using perplexity, or loss, as a measure of sample value for data selection guidance fails to account for diversity and difficulty, resulting in considerably poorer performance. The RAND strategy, which can be seen as a simple approach that focuses solely on diversity to capture the overall data distribution by random sampling, achieves performance slightly weaker than Full. However, it falls short of outperforming Full due to the lack of more in-depth sample-level evaluation. The IFD method provides a deeper analysis of sample-wise instruction-following difficulty, helping the model enhance performance with less data. However, as it relies solely on a single metric for data selection, it leaves room for further improvement. Besides, the results reported by AlpacaEval show that it is slightly less competitive than the Full strategy on two high-quality instruction tuning datasets. The performance comparisons on the real-world dataset are presented in Table~\ref{tab:exp.1.1}, under a fixed selection budget $k=10\%$, further demonstrating the superiority of our method.

\begin{table}[t]
\small
\centering
\caption{Comparisons of D\textsubscript{3} with three ablation strategies. Scores represent the average winning scores across all test sets.}
\label{tab:exp.3}
\begin{tabular}{@{}l|ccc@{}}
\toprule
 & \multicolumn{3}{c}{\textbf{Winning Score} (\textit{5\%} vs. \textbf{Full})} \\ \midrule
\textbf{Strategy} & Alpaca & AlpacaGPT4 & WizardLM \\ \midrule
\textbf{D\textsubscript{3}-w/o1} & 0.38 & 0.83 & 0.85 \\
\textbf{D\textsubscript{3}-w/o2} & 1.08 & 1.03 & 1.01 \\
\textbf{D\textsubscript{3}-w/o3} & 1.12 & 1.12 & 1.06 \\
\midrule
\textbf{D\textsubscript{3}} & \textbf{1.27} & \textbf{1.19} & \textbf{1.11} \\ \bottomrule
\end{tabular}
\vspace{-7pt}
\end{table}

\subsection{Performance with Varied Budgets}
We systematically examine the impact of selection budgets on various strategies by incrementally increasing the budget. In line with the setup in the main experiment, all strategies select the required data in a single round. Figure~\ref{fig:exp2} presents the winning score on the Alpaca dataset. As the budget increases, the performance of more potent strategies exhibits a trend of initially improving and then declining. Conversely, weaker strategies gradually approach the performance of the Full strategy. This is because selecting more valuable data positively contributes to model instruction tuning. However, once the effective data reaches saturation, incorporating additional data of slightly lower quality or with redundancy leads to diminishing returns, ultimately causing the performance to converge with that of the Full strategy.
Specifically, D\textsubscript{3} achieves superior performance with the smallest budget, highlighting the effectiveness of our strategy in selecting the most valuable coreset data for sample-efficient instruction tuning. Similarly, the results on the other two high-quality datasets, as reported in the appendix, also empirically validate this.
In contrast, other strategies display weaker performance trajectories or slower growth, which further underscores the superiority and effectiveness of our approach.

\subsection{Ablation Study}
We conducted the following ablation experiments to validate the effectiveness of the three data selection criteria in D\textsubscript{3}.
\begin{itemize}
    \item \textbf{D\textsubscript{3}-w/o1}: Removes diversity.
    \item \textbf{D\textsubscript{3}-w/o2}: Removes difficulty.
    \item \textbf{D\textsubscript{3}-w/o3}: Removes dependability.
\end{itemize}

Table~\ref{tab:exp.3} shows the performance of the three ablation strategies. Given that Alpaca is a medium-quality dataset, neglecting diversity causes the model to overly focus on certain challenging and highly dependable samples, resulting in a significant performance drop. In the case of the other two high-quality datasets, most samples have dependability scores close to 1, making D\textsubscript{3}-w/o2, to some extent, a strategy that only considers diversity. As a result, its performance is similar to but slightly better than the RAND strategy shown in the main results. The performance impact of removing dependability is more pronounced for Alpaca due to the quality difference of datasets. These validate the necessity of comprehensively considering all three criteria in data selection.

\subsection{Selection with Multi-Round}
\begin{figure}[t]
    \centering
    \vspace{-10pt}
    \includegraphics[width=\columnwidth]{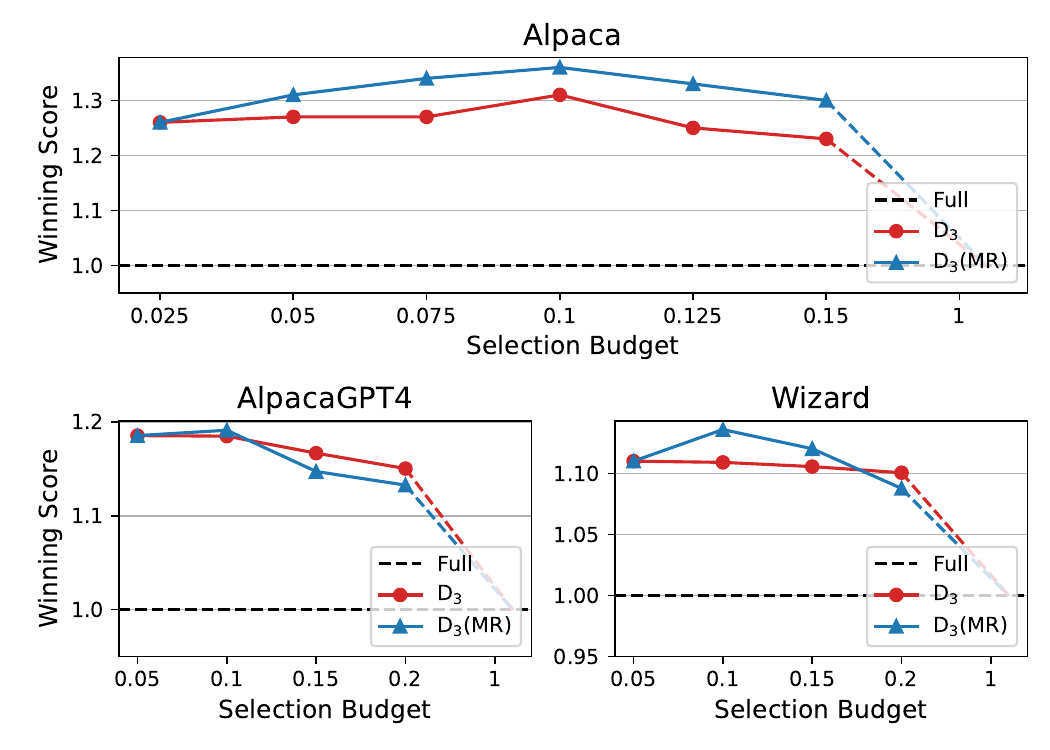}
    \vspace{-15pt}
    \caption{The performance comparisons of D\textsubscript{3} and D\textsubscript{3}(MR), which selects data through multi-round scoring and selection.}
    \label{fig:exp4}
    \vspace{-10pt}
\end{figure}

In this section, we present the performance of the D\textsubscript{3} with multi-round selection. The budget increment is 2.5\% for the Alpaca dataset and 5\% for the other two datasets.  Figure~\ref{fig:exp4} compares the performance of multi-round selection D\textsubscript{3}(MR) with single-round selection D\textsubscript{3}.
As observed, on the Alpaca dataset with medium quality, D\textsubscript{3}(MR) could adaptively refine the selection focus during the instruction tuning process by leveraging real-time model feedback, thereby better identifying the most valuable subset of data to enhance performance significantly. On datasets with inherently high-quality samples, although the selected small amount of data is already sufficient, multi-round selection still provides some benefits in rapidly improving performance. Note that since D\textsubscript{3}(MR) involves re-scoring in each round, employing multi-round selection necessitates a trade-off between performance gains and the additional computational overhead.

\section{Conclusion}

This paper investigates data selection for sample-efficient LLM instruction tuning. We first establish data selection criteria based on three distinct aspects of diversity, difficulty, and dependability, offering the systematic guidance for data selection. We then propose a novel method D\textsubscript{3} to identify valuable subsets from large datasets by scoring and jointly optimizing three criteria. Our results show that instruction-following capabilities can be achieved with significantly smaller datasets, improving the effectiveness and efficiency of instruction tuning. One direction for future work is to explore token-wise weights on loss based on selection during instruction tuning.

\clearpage

\appendix

\onecolumn
\section{Appendix}
This appendix provides supplementary details to support the main text, including further descriptions of the methodology and additional implementation details and results in experiments.

\subsection{Methodological Supplement}
\setcounter{subsubsection}{0}
\subsubsection{Sample Diversity}
As described in the main text, the sample diversity function we defined depends on the calculation between the embeddings of different samples. 
The sample embeddings are obtained from the LLM to be fine-tuned, rather than introducing external models, thereby providing an unbiased representation of the relationships between samples within the current LLM's feature space. Due to the decoder-based transformer architecture of current mainstream LLMs, it is hard to directly obtain a semantic embedding for each sample. Therefore, as shown in Algorithm~\ref{alg:embeddings}, we average the final-layer hidden features of each token in the sample to obtain a vector as its embedding.

\begin{algorithm}[h!]
\caption{The Sample Embedding Calculation.}
\label{alg:embeddings}
\begin{algorithmic}[1]
    \STATE \textbf{Input:} Sample $z=\{x,y\}$, LLM $F$.
    \STATE \textbf{Procedure:}
    \STATE $\hat{x}, \hat{y} \leftarrow \mathrm{tokenization(x, y)}$
    \FOR{$t := 0 \rightarrow |\hat{y}|$}
        \STATE $\mathrm{hidden}_t\leftarrow F_\mathrm{last\_hidden}(\hat{x}, \hat{y}_{<t})$
    \ENDFOR
    \STATE $\mathrm{embedding}\leftarrow \frac{1}{|\hat{y}|} \sum_{t=0}^{|\hat{y}|}\mathrm{hidden}_t$
    \RETURN $\mathrm{embedding}$
\end{algorithmic}
\end{algorithm}

\subsubsection{Sample Dependability}
We introduce an external teacher LLM to evaluate sample-model dependability. We generate a prompt for each sample and input it to the teacher model to obtain an assessment of the sample’s dependability. Specifically, the prompt template, as shown in Figure~\ref{fig.prompt1}, presents dependability assessment criteria in the form of instructions. These criteria can be tailored to different tasks by customizing specific evaluation preferences. The teacher model then outputs an evaluation score of 0 (bad) or 1 (good) on a new line, corresponding to negative and positive tokens $w_{neg}$ and $ w_{pos}$, respectively. For evaluation efficiency, we limit the maximum number of tokens generated by the teacher model to 1 and truncate subsequent explanatory content related to the score. We compute the soft scores as sample dependability using the logits output by the teacher model on these two tokens.

\begin{figure}[b!]
\small
\caption{The prompt template used to evaluation the sample depenability.}
\centering
\label{fig.prompt1}
\begin{tabular}{@{}l@{}}
\toprule \midrule
\textbf{Sample Dependability Evaluation Prompt} \\ \midrule
You are tasked with evaluating the sample based on whether it meets all the following criteria: \\
- Fluency: Is the {[}Response{]} coherent and free from irrelevant content or nonsensical marks? \\
- Accuracy: Does the {[}Response{]} correctly answer the {[}Query{]} without providing false information? \\
- Clarity: Is the {[}Response{]} clear and logically structured? \\
""" \\
{[}Query{]} \\
\{instruction\} \\
{[}Response{]} \\
\{response\} \\
""" \\
Please review the sample and assign a quality score of 0 (bad) or 1 (good). \\ \midrule
Quality score (0 or 1): \\ \midrule \bottomrule
\end{tabular}
\end{figure}

\begin{equation}
d_3(z)=\frac{\exp(F^*(w_{pos}\mid T(z)))}{\exp(F^*(w_{pos}\mid T(z)))+\exp(F^*(w_{neg}\mid T(z)))},
\end{equation}
where $ F^* $ represents the teacher model, $ T(z) $ is the evaluation prompt generated based on sample $ z $, and $ F^*(w | T) $ refers to the logit of the next token $ w $ given the inputs $ T(z) $.

\subsection{Experimental Supplement}
\subsubsection{Evaluation Details}
In experiments, we employ two evaluation methods: pairwise evaluation and the AlpacaEval Leaderboard~\cite{alpaca_eval}. The pairwise evaluation employs a more powerful model as a judge to assess the quality of two different responses to the same instruction. Specifically, we construct the pairwise evaluation prompt as shown in Figure~\ref{fig.prompt2} and pass it to the judge LLM for evaluation. The judge model used for evaluation inference in our experiments is the locally deployable Qwen2.5-32B-Instruct~\cite{qwen2.5,qwen2}, which possesses extensive knowledge and strong instruction-following capabilities. The test datasets involved in pairwise evalutation are WizardLM~\cite{xu2023wizardlm}, Self-instruct~\cite{wang2023selfinstruct}, Vicuna\cite{vicuna2023}, Koala~\cite{vu2023koala}, and LIMA\cite{zhou2023lima}. They consist of 218, 252, 80, 180, 300 human-curated instructions, respectively, spanning domains such as mathematics, coding, writing, knowledge, and computer science.
For the AlpacaEval Leaderboard, we deployed the official AlpacaEval repository\footnote{https://github.com/tatsu-lab/alpaca\_eval} locally and used the GPT-4o API for evaluation. Additionally, the widely used Open LLM Leaderboard benchmark could not be deployed locally for evaluation in this experiment due to the absence of certain original test datasets in the repository. Given the large scale and high credibility of the two evaluation methods described above, this leaderboard was temporarily excluded but may be incorporated in future evaluations.

\begin{figure}[t!]
\centering
\caption{The prompt template used for pairwise evaluation of the model response quality.}
\label{fig.prompt2}
\begin{tabular}{@{}p{\textwidth}@{}}
\toprule \midrule
\textbf{Pairwise Evaluation Prompt for Response} \\ \midrule
\textbf{System Prompt} \\
You are a helpful and precise assistant for checking the quality of the answer. \\
\textbf{User Prompt} \\
{[}Question{]} \\
\{instruction\} \\
{[}The Start of Assistant 1's Answer{]} \\
\{response 1\} \\
{[}The End of Assistant 1's Answer{]} \\
{[}The Start of Assistant 2's Answer{]} \\
\{response 2\} \\
{[}The End of Assistant 2's Answer{]} \\
We would like to request your feedback on the performance of two AI assistants in response to the user question displayed above. \\
Please rate the helpfulness, relevance, accuracy, level of details of their responses. Each assistant receives an overall score on a scale of 1 to 10, where a higher score indicates better overall performance. Please first output a single line containing only two values indicating the scores for Assistant 1 and 2, respectively. The two scores are separated by a space. In the subsequent line, please provide a comprehensive explanation of your evaluation, avoiding any potential bias and ensuring that the order in which the responses were presented does not affect your judgment. \\ \midrule \bottomrule
\end{tabular}
\end{figure}

\subsubsection{Implementation Details}
In experiments, we use the pre-trained Llama2-7B~\cite{touvron2023llamaa} as the initial model. All methods begin with two training epochs on a warm-up set, comprising 1\% of the data randomly sampled from the data pool. Next, each strategy selects a data subset of equal size to reach the given selection budget for instruction tuning. The selected samples are then trained for three epochs from the warm-up model to obtain the final fine-tuned LLMs. The model training and inference are implemented based on the Llama-Factory~\cite{zheng2024llamafactory} framework. We use the AdamW optimizer with a batch size of 128. The maximum input length for model training is 2048, while the maximum output length for inference is 512. The initial learning rate for the warm-up step is set to 1e-5, with a starting rate of 2e-5 for training on selected data. A cosine annealing learning rate schedule is applied during the process. In the multi-round selection of D\textsubscript{3}(MR), each round involves training for 2 epochs on the selected data in last round. Finally, the warm-up model is trained using the same setting as described above, based on all the selected data from each round, to obtain the final model. The sample dependability is calculated using the Qwen2.5-14B-Instruct~\cite{qwen2,qwen2.5} model. For the real-world dataset, we use Qwen2.5-7B as the initial model, with other settings consistent with the above discussion.

\subsubsection{Additional Experimental Results}
In this section, we provide the detailed results for each experiment.
\subsubsection{Performance with Varied Budgets}
The results with varied budgets of different strategies are presented in Table~\ref{tab:1}, \ref{tab:2} and ~\ref{tab:3} for the Alpaca, AlpacaGPT4 and WizardLM datasets, respectively.

\subsubsection{Ablation Study}
The detailed comparison results of the ablation strategies are provided in Table~\ref{tab:4}.

\subsubsection{Selection with Multi-Round}
The detailed comparison results of D\textsubscript{3} and D\textsubscript{3}(MR) are presented in Table~\ref{tab:5}.

\begin{table}[ht]
\centering
\small
\caption{Performance with varied selection budgets on the Alpaca dataset.}
\label{tab:1}
\begin{tabular}{@{}cc|cccccc|c@{}}
\toprule \midrule
                                                                          &                 & \multicolumn{6}{c|}{\textbf{Winning Score (vs Full)}}                       & \textbf{Leaderboard} \\ \midrule
\multicolumn{1}{c|}{\textbf{Strategy}}                                    & \textbf{Budget} & WizardLM & Sinstruct & Vicuna & Koala & LIMA                      & Average & AlpacaEval           \\ \midrule

\multicolumn{1}{c|}{Full}                                                 & 1.000           & 1.00     & 1.00      & 1.00   & 1.00  & \multicolumn{1}{c|}{1.00} & \multicolumn{1}{c|}{1.00}    & 28.00                \\ \midrule \midrule
\multicolumn{1}{c|}{\multirow{6}{*}{D\textsubscript{3}}} & 0.025           & 1.22     & 1.14      & 1.33   & 1.26  & \multicolumn{1}{c|}{1.38} & \multicolumn{1}{c|}{1.26}    & 40.49                \\
\multicolumn{1}{c|}{}                                                     & 0.050           & 1.22     & 1.16      & 1.21   & 1.33  & \multicolumn{1}{c|}{1.37} & \multicolumn{1}{c|}{1.27}    & 42.01                \\
\multicolumn{1}{c|}{}                                                     & 0.075           & 1.31     & 1.20      & 1.23   & 1.21  & \multicolumn{1}{c|}{1.34} & \multicolumn{1}{c|}{1.27}    & 43.62                \\
\multicolumn{1}{c|}{}                                                     & 0.100           & 1.29     & 1.23      & 1.37   & 1.33  & \multicolumn{1}{c|}{1.36} & \multicolumn{1}{c|}{1.31}    & 43.12                \\
\multicolumn{1}{c|}{}                                                     & 0.125           & 1.25     & 1.17      & 1.18   & 1.28  & \multicolumn{1}{c|}{1.31} & \multicolumn{1}{c|}{1.25}    & 42.06                \\
\multicolumn{1}{c|}{}                                                     & 0.150           & 1.21     & 1.21      & 1.17   & 1.24  & \multicolumn{1}{c|}{1.27} & \multicolumn{1}{c|}{1.23}    & 40.04                \\ \midrule \midrule
\multicolumn{1}{c|}{\multirow{6}{*}{PPL}}                                 & 0.025           & 0.11     & 0.19      & 0.04   & 0.13  & \multicolumn{1}{c|}{0.07} & \multicolumn{1}{c|}{0.12}    & -                    \\
\multicolumn{1}{c|}{}                                                     & 0.050           & 0.20     & 0.30      & 0.08   & 0.16  & \multicolumn{1}{c|}{0.15} & \multicolumn{1}{c|}{0.19}    & 4.34                 \\
\multicolumn{1}{c|}{}                                                     & 0.075           & 0.29     & 0.38      & 0.09   & 0.28  & \multicolumn{1}{c|}{0.17} & \multicolumn{1}{c|}{0.26}    & -                    \\
\multicolumn{1}{c|}{}                                                     & 0.100           & 0.42     & 0.46      & 0.19   & 0.43  & \multicolumn{1}{c|}{0.33} & \multicolumn{1}{c|}{0.39}    & 9.27                 \\
\multicolumn{1}{c|}{}                                                     & 0.125           & 0.63     & 0.61      & 0.65   & 0.61  & \multicolumn{1}{c|}{0.48} & \multicolumn{1}{c|}{0.58}    & \textbf{-}           \\
\multicolumn{1}{c|}{}                                                     & 0.150           & 0.71     & 0.6       & 0.64   & 0.66  & \multicolumn{1}{c|}{0.68} & \multicolumn{1}{c|}{0.66}    & -                    \\ \midrule \midrule
\multicolumn{1}{c|}{\multirow{6}{*}{Random}}                              & 0.025           & 0.84     & 0.79      & 0.91   & 0.90  & \multicolumn{1}{c|}{0.93} & \multicolumn{1}{c|}{0.87}    & 26.69                \\
\multicolumn{1}{c|}{}                                                     & 0.050           & 0.99     & 0.93      & 1.09   & 0.98  & \multicolumn{1}{c|}{0.95} & \multicolumn{1}{c|}{0.97}    & 28.29                \\
\multicolumn{1}{c|}{}                                                     & 0.075           & 0.95     & 0.90      & 0.99   & 0.95  & \multicolumn{1}{c|}{0.99} & \multicolumn{1}{c|}{0.95}    & 28.09                \\
\multicolumn{1}{c|}{}                                                     & 0.100           & 0.88     & 0.96      & 0.91   & 0.94  & \multicolumn{1}{c|}{0.93} & \multicolumn{1}{c|}{0.93}    & 28.24                \\
\multicolumn{1}{c|}{}                                                     & 0.125           & 0.98     & 0.89      & 1.03   & 0.98  & \multicolumn{1}{c|}{1.03} & \multicolumn{1}{c|}{0.98}    & 29.87                \\
\multicolumn{1}{c|}{}                                                     & 0.150           & 0.98     & 0.98      & 0.94   & 1.01  & \multicolumn{1}{c|}{1.04} & \multicolumn{1}{c|}{1.00}    & 30.35                \\ \midrule \midrule
\multicolumn{1}{c|}{\multirow{6}{*}{IFD}}                                 & 0.025           & 1.13     & 0.89      & 1.48   & 1.1   & \multicolumn{1}{c|}{1.28} & \multicolumn{1}{c|}{1.14}    & -                    \\
\multicolumn{1}{c|}{}                                                     & 0.050           & 1.15     & 1.01      & 1.38   & 1.13  & \multicolumn{1}{c|}{1.28} & \multicolumn{1}{c|}{1.17}    & 36.50                 \\
\multicolumn{1}{c|}{}                                                     & 0.075           & 1.08     & 1.14      & 1.49   & 1.24  & \multicolumn{1}{c|}{1.30}  & \multicolumn{1}{c|}{1.22}    & -                    \\
\multicolumn{1}{c|}{}                                                     & 0.100           & 1.14     & 1.11      & 1.46   & 1.27  & \multicolumn{1}{c|}{1.27} & \multicolumn{1}{c|}{1.22}    & 39.49                \\
\multicolumn{1}{c|}{}                                                     & 0.125           & 1.22     & 1.13      & 1.46   & 1.25  & \multicolumn{1}{c|}{1.25} & \multicolumn{1}{c|}{1.23}    & -                    \\
\multicolumn{1}{c|}{}                                                     & 0.150           & 1.11     & 1.12      & 1.35   & 1.16  & \multicolumn{1}{c|}{1.23} & \multicolumn{1}{c|}{1.17}    & -                    \\ \midrule\bottomrule
\end{tabular}
\end{table}

\begin{table}[ht]
\centering
\small
\caption{Performance with varied selection budgets on the AlpacaGPT4 dataset.}
\label{tab:2}
\begin{tabular}{@{}cc|cccccc|c@{}}
\toprule \midrule
                                                                          &                 & \multicolumn{6}{c|}{\textbf{Winning Score (vs Full)}}                       & \textbf{Leaderboard} \\ \midrule
\multicolumn{1}{c|}{\textbf{Strategy}}                                    & \textbf{Budget} & WizardLM & Sinstruct & Vicuna & Koala & LIMA                      & Average & AlpacaEval           \\ \midrule
\multicolumn{1}{c|}{Full}                                                 & 1.00            & 1.00     & 1.00      & 1.00   & 1.00  & \multicolumn{1}{c|}{1.00} & 1.00    & 65.68                \\ \midrule \midrule
\multicolumn{1}{c|}{\multirow{4}{*}{D\textsubscript{3}}} & 0.05            & 1.13     & 1.19      & 1.07   & 1.24  & \multicolumn{1}{c|}{1.22} & 1.19    & 71.26                \\
\multicolumn{1}{c|}{}                                                     & 0.10            & 1.11     & 1.20      & 1.06   & 1.28  & \multicolumn{1}{c|}{1.22} & 1.19    & 72.02                \\
\multicolumn{1}{c|}{}                                                     & 0.15            & 1.09     & 1.23      & 1.06   & 1.18  & \multicolumn{1}{c|}{1.19} & 1.17    & 68.28                \\
\multicolumn{1}{c|}{}                                                     & 0.20            & 1.09     & 1.12      & 1.06   & 1.24  & \multicolumn{1}{c|}{1.19} & 1.15    & 71.27                \\ \midrule \midrule
\multicolumn{1}{c|}{\multirow{4}{*}{PPL}}                                 & 0.05            & 0.29     & 0.33      & 0.11   & 0.36  & \multicolumn{1}{c|}{0.20} & 0.27    & 20.92                \\
\multicolumn{1}{c|}{}                                                     & 0.10            & 0.62     & 0.69      & 0.51   & 0.65  & \multicolumn{1}{c|}{0.66} & 0.65    & -                    \\
\multicolumn{1}{c|}{}                                                     & 0.15            & 0.70     & 0.66      & 0.68   & 0.62  & \multicolumn{1}{c|}{0.74} & 0.69    & -                    \\
\multicolumn{1}{c|}{}                                                     & 0.20            & 0.77     & 0.79      & 0.71   & 0.87  & \multicolumn{1}{c|}{0.87} & 0.82    & -                    \\ \midrule \midrule
\multicolumn{1}{c|}{\multirow{4}{*}{Random}}                              & 0.05            & 0.85     & 0.86      & 0.83   & 0.92  & \multicolumn{1}{c|}{0.98} & 0.90    & 60.93                \\
\multicolumn{1}{c|}{}                                                     & 0.10            & 0.89     & 1.05      & 0.99   & 1.00  & \multicolumn{1}{c|}{0.86} & 0.95    & -                    \\
\multicolumn{1}{c|}{}                                                     & 0.15            & 0.89     & 1.02      & 0.91   & 1.04  & \multicolumn{1}{c|}{0.95} & 0.97    & -                    \\
\multicolumn{1}{c|}{}                                                     & 0.20            & 0.88     & 0.93      & 0.89   & 1.08  & \multicolumn{1}{c|}{0.94} & 0.95    & -                    \\ \midrule \midrule
\multicolumn{1}{c|}{\multirow{4}{*}{IFD}}                                 & 0.05            & 1.10     & 1.08      & 1.00   & 1.11  & \multicolumn{1}{c|}{1.15} & 1.10    & 64.53                \\
\multicolumn{1}{c|}{}                                                     & 0.10            & 1.08     & 1.17      & 1.06   & 1.15  & \multicolumn{1}{c|}{1.25} & 1.16    & -                    \\
\multicolumn{1}{c|}{}                                                     & 0.15            & 1.08     & 1.13      & 1.17   & 1.19  & \multicolumn{1}{c|}{1.29} & 1.18    & -                    \\
\multicolumn{1}{c|}{}                                                     & 0.20            & 1.11     & 1.16      & 1.18   & 1.21  & \multicolumn{1}{c|}{1.22} & 1.18    & -                    \\ \midrule \bottomrule
\end{tabular}
\end{table}

\begin{table}[ht]
\centering
\small
\caption{Performance with varied selection budgets on the WizardLM dataset.}
\label{tab:3}
\begin{tabular}{@{}cc|cccccc|c@{}}
\toprule \midrule
                                                                          &                 & \multicolumn{6}{c|}{\textbf{Winning Score (vs Full)}}                       & \textbf{Leaderboard} \\ \midrule
\multicolumn{1}{c|}{\textbf{Strategy}}                                    & \textbf{Budget} & WizardLM & Sinstruct & Vicuna & Koala & LIMA                      & Average & AlpacaEval           \\ \midrule
\multicolumn{1}{c|}{Full}                                                 & 1.00            & 1.00     & 1.00      & 1.00   & 1.00  & \multicolumn{1}{c|}{1.00} & 1.00    & 58.42                \\ \midrule \midrule
\multicolumn{1}{c|}{\multirow{4}{*}{D\textsubscript{3}}} & 0.05            & 0.99     & 1.14      & 1.21   & 1.07  & \multicolumn{1}{c|}{1.17} & 1.11    & 63.09                \\
\multicolumn{1}{c|}{}                                                     & 0.10            & 1.02     & 1.16      & 1.15   & 1.11  & \multicolumn{1}{c|}{1.12} & 1.11    & 63.28                \\
\multicolumn{1}{c|}{}                                                     & 0.15            & 0.97     & 1.12      & 1.24   & 1.10  & \multicolumn{1}{c|}{1.16} & 1.11    & 61.00                \\
\multicolumn{1}{c|}{}                                                     & 0.20            & 0.96     & 1.12      & 1.24   & 1.15  & \multicolumn{1}{c|}{1.12} & 1.10    & 62.21                \\ \midrule \midrule
\multicolumn{1}{c|}{\multirow{4}{*}{PPL}}                                 & 0.05            & 0.40     & 0.52      & 0.60   & 0.45  & \multicolumn{1}{c|}{0.48} & 0.48    & 28.49                \\
\multicolumn{1}{c|}{}                                                     & 0.10            & 0.63     & 0.71      & 0.94   & 0.66  & \multicolumn{1}{c|}{0.74} & 0.71    & -                    \\
\multicolumn{1}{c|}{}                                                     & 0.15            & 0.62     & 0.75      & 0.93   & 0.79  & \multicolumn{1}{c|}{0.73} & 0.74    & -                    \\
\multicolumn{1}{c|}{}                                                     & 0.20            & 0.74     & 0.80      & 1.06   & 0.84  & \multicolumn{1}{c|}{0.82} & 0.82    & -                    \\ \midrule \midrule
\multicolumn{1}{c|}{\multirow{4}{*}{Random}}                              & 0.05            & 0.92     & 0.81      & 1.08   & 0.79  & \multicolumn{1}{c|}{0.94} & 0.89    & 54.48                \\
\multicolumn{1}{c|}{}                                                     & 0.10            & 0.93     & 0.87      & 1.13   & 0.98  & \multicolumn{1}{c|}{0.91} & 0.93    & -                    \\
\multicolumn{1}{c|}{}                                                     & 0.15            & 0.91     & 0.90      & 1.01   & 0.93  & \multicolumn{1}{c|}{0.93} & 0.92    & -                    \\
\multicolumn{1}{c|}{}                                                     & 0.20            & 0.92     & 0.95      & 1.14   & 0.98  & \multicolumn{1}{c|}{0.95} & 0.96    & -                    \\ \midrule \midrule
\multicolumn{1}{c|}{\multirow{4}{*}{IFD}}                                 & 0.05            & 0.93     & 0.93      & 1.28   & 0.99  & \multicolumn{1}{c|}{1.11} & 1.02    & 54.94                \\
\multicolumn{1}{c|}{}                                                     & 0.10            & 0.97     & 1.02      & 1.17   & 1.03  & \multicolumn{1}{c|}{1.06} & 1.03    & -                    \\
\multicolumn{1}{c|}{}                                                     & 0.15            & 1.01     & 0.98      & 1.18   & 1.04  & \multicolumn{1}{c|}{1.10} & 1.05    & -                    \\
\multicolumn{1}{c|}{}                                                     & 0.20            & 0.97     & 1.05      & 1.24   & 1.10  & \multicolumn{1}{c|}{1.06} & 1.06    & -                    \\ \midrule \bottomrule
\end{tabular}
\end{table}

\begin{table}[ht]
\centering
\footnotesize
\caption{Detailed results of ablation strategies.}
\label{tab:4}
\begin{tabular}{@{}cl|cccccc@{}}
\toprule \midrule
\multicolumn{1}{l}{}                                                                             &                   & \multicolumn{6}{c}{\textbf{Winning Score (vs. Full)}}                                                              \\ \midrule
\multicolumn{1}{c|}{\textbf{Dataset}}                                                            & \textbf{Strategy} & WizardLM (218)      & Sinstruct (252)     & Vicuna (80)       & Koala (180)        & \multicolumn{1}{c|}{LIMA (300)}          & Average       \\ \midrule
\multicolumn{1}{c|}{\multirow{4}{*}{\begin{tabular}[c]{@{}c@{}}Alpaca\\ (5\%)\end{tabular}}}     & \textbf{D\textsubscript{3}-w/o1}  & 0.40          & 0.52          & 0.21          & 0.35          & \multicolumn{1}{c|}{0.30}          & 0.38          \\
\multicolumn{1}{c|}{}                                                                            & \textbf{D\textsubscript{3}-w/o2}  & 1.10          & 1.09          & 1.02          & 1.07          & \multicolumn{1}{c|}{1.09}          & 1.08          \\
\multicolumn{1}{c|}{}                                                                            & \textbf{D\textsubscript{3}-w/o3}  & 1.10          & 1.04          & 1.10          & 1.14          & \multicolumn{1}{c|}{1.19}          & 1.12          \\
\multicolumn{1}{c|}{}                                                                            & \textbf{D\textsubscript{3}}       & 1.22          & 1.16          & 1.21          & 1.33          & \multicolumn{1}{c|}{1.37}          & 1.27          \\ \midrule \midrule
\multicolumn{1}{c|}{\multirow{4}{*}{\begin{tabular}[c]{@{}c@{}}AlpacaGPT4\\ (5\%)\end{tabular}}} & \textbf{D\textsubscript{3}-w/o1}  & 0.84          & 0.82          & 0.79          & 0.84          & \multicolumn{1}{c|}{0.82}          & 0.83          \\
\multicolumn{1}{c|}{}                                                                            & \textbf{D\textsubscript{3}-w/o2}  & 0.99          & 1.07          & 0.99          & 1.02          & \multicolumn{1}{c|}{1.03}          & 1.03          \\
\multicolumn{1}{c|}{}                                                                            & \textbf{D\textsubscript{3}-w/o3}  & 1.11          & 1.13          & 1.14          & 1.10          & \multicolumn{1}{c|}{1.13}          & 1.12          \\
\multicolumn{1}{c|}{}                                                                            & \textbf{D\textsubscript{3}}       & 1.13          & 1.19          & 1.07          & 1.24          & \multicolumn{1}{c|}{1.22}          & 1.17          \\ \midrule \midrule
\multicolumn{1}{c|}{\multirow{4}{*}{\begin{tabular}[c]{@{}c@{}}WizardLM\\ (5\%)\end{tabular}}}   & \textbf{D\textsubscript{3}-w/o1}  & 0.76 & 0.85          & 0.92          & 0.81          & \multicolumn{1}{c|}{0.93}          & 0.85          \\
\multicolumn{1}{c|}{}                                                                            & \textbf{D\textsubscript{3}-w/o2}  & 0.99          & 1.02          & 1.03          & 1.02          & \multicolumn{1}{c|}{1.01}          & 1.01          \\
\multicolumn{1}{c|}{}                                                                            & \textbf{D\textsubscript{3}-w/o3}  & 1.05 & 1.01          & 1.20          & 1.01          & \multicolumn{1}{c|}{1.09}          & 1.06          \\
\multicolumn{1}{c|}{}                                                                            & \textbf{D\textsubscript{3}}       & 0.99          & 1.14 & 1.21 & 1.07 & \multicolumn{1}{c|}{1.17} & 1.11 \\ \midrule \bottomrule
\end{tabular}
\end{table}

\begin{table}[ht]
\scriptsize
\centering
\caption{Detailed comparision result of D\textsubscript{3} and D\textsubscript{3}(MR) for three instruction tuning datasets.}
\label{tab:5}
\begin{tabular}{@{}cc|cccccccccccc|cc@{}}
\toprule \midrule
                                                                                            &                 & \multicolumn{12}{c|}{\textbf{Winning Score (vs. Full)}}                                                                                                                                                                                                                                                                    & \multicolumn{2}{c}{\textbf{Leaderboard}} \\ \cmidrule(l){3-16} 
\textbf{}                                                                                   & \textbf{}       & \multicolumn{2}{c|}{WizardLM}                         & \multicolumn{2}{c|}{Sinstruct}                        & \multicolumn{2}{c|}{Vicuna}                           & \multicolumn{2}{c|}{Koala}                            & \multicolumn{2}{c|}{LIMA}                             & \multicolumn{2}{c|}{Average}       & \multicolumn{2}{c}{AlpacaEval}           \\ \midrule
\textbf{Dataset}                                                                            & \textbf{Budget} & \multicolumn{1}{c|}{D\textsubscript{3}}   & \multicolumn{1}{c|}{MR}   & \multicolumn{1}{c|}{D\textsubscript{3}}   & \multicolumn{1}{c|}{MR}   & \multicolumn{1}{c|}{D\textsubscript{3}}   & \multicolumn{1}{c|}{MR}   & \multicolumn{1}{c|}{D\textsubscript{3}}   & \multicolumn{1}{c|}{MR}   & \multicolumn{1}{c|}{D\textsubscript{3}}   & \multicolumn{1}{c|}{MR}   & \multicolumn{1}{c|}{D\textsubscript{3}}    & MR    & \multicolumn{1}{c|}{D\textsubscript{3}}       & MR       \\ \midrule \midrule
\multicolumn{1}{c|}{\multirow{6}{*}{Alpaca}}                                                & 0.025           & 1.22                      & \multicolumn{1}{c|}{1.22} & 1.14                      & \multicolumn{1}{c|}{1.14} & 1.33                      & \multicolumn{1}{c|}{1.33} & 1.26                      & \multicolumn{1}{c|}{1.26} & 1.38                      & \multicolumn{1}{c|}{1.38} & 40.49                      & 40.49 & 40.49                         & 40.49    \\
\multicolumn{1}{c|}{}                                                                       & 0.050           & \multicolumn{1}{c|}{1.22} & \multicolumn{1}{c|}{1.27} & \multicolumn{1}{c|}{1.16} & \multicolumn{1}{c|}{1.16} & \multicolumn{1}{c|}{1.21} & \multicolumn{1}{c|}{1.38} & \multicolumn{1}{c|}{1.33} & \multicolumn{1}{c|}{1.34} & \multicolumn{1}{c|}{1.37} & \multicolumn{1}{c|}{1.43} & \multicolumn{1}{c|}{42.01} & 42.88 & \multicolumn{1}{c|}{42.01}    & 42.88    \\
\multicolumn{1}{c|}{}                                                                       & 0.075           & \multicolumn{1}{c|}{1.31} & \multicolumn{1}{c|}{1.33} & \multicolumn{1}{c|}{1.20} & \multicolumn{1}{c|}{1.25} & \multicolumn{1}{c|}{1.23} & \multicolumn{1}{c|}{1.34} & \multicolumn{1}{c|}{1.21} & \multicolumn{1}{c|}{1.34} & \multicolumn{1}{c|}{1.34} & \multicolumn{1}{c|}{1.43} & \multicolumn{1}{c|}{43.62} & 44.12 & \multicolumn{1}{c|}{43.62}    & 44.12    \\
\multicolumn{1}{c|}{}                                                                       & 0.100           & \multicolumn{1}{c|}{1.29} & \multicolumn{1}{c|}{1.42} & \multicolumn{1}{c|}{1.23} & \multicolumn{1}{c|}{1.25} & \multicolumn{1}{c|}{1.37} & \multicolumn{1}{c|}{1.44} & \multicolumn{1}{c|}{1.33} & \multicolumn{1}{c|}{1.37} & \multicolumn{1}{c|}{1.36} & \multicolumn{1}{c|}{1.38} & \multicolumn{1}{c|}{43.12} & 45.09 & \multicolumn{1}{c|}{43.12}    & 45.09    \\
\multicolumn{1}{c|}{}                                                                       & 0.125           & \multicolumn{1}{c|}{1.25} & \multicolumn{1}{c|}{1.31} & \multicolumn{1}{c|}{1.17} & \multicolumn{1}{c|}{1.23} & \multicolumn{1}{c|}{1.18} & \multicolumn{1}{c|}{1.39} & \multicolumn{1}{c|}{1.28} & \multicolumn{1}{c|}{1.39} & \multicolumn{1}{c|}{1.31} & \multicolumn{1}{c|}{1.36} & \multicolumn{1}{c|}{42.06} & 40.71 & \multicolumn{1}{c|}{42.06}    & 40.71    \\
\multicolumn{1}{c|}{}                                                                       & 0.150           & 1.21                      & \multicolumn{1}{c|}{1.29} & 1.21                      & \multicolumn{1}{c|}{1.25} & 1.17                      & \multicolumn{1}{c|}{1.27} & 1.24                      & \multicolumn{1}{c|}{1.31} & 1.27                      & \multicolumn{1}{c|}{1.34} & 40.04                      & 44.08 & 40.04                         & 44.08    \\ \midrule \midrule
\multicolumn{1}{c|}{\multirow{4}{*}{\begin{tabular}[c]{@{}c@{}}Alpaca\\ GPT4\end{tabular}}} & 0.05            & 1.13                      & \multicolumn{1}{c|}{1.13} & 1.19                      & \multicolumn{1}{c|}{1.19} & 1.07                      & \multicolumn{1}{c|}{1.07} & 1.24                      & \multicolumn{1}{c|}{1.24} & 1.22                      & \multicolumn{1}{c|}{1.22} & 1.19                       & 1.19  & 71.26                         & 71.26    \\
\multicolumn{1}{c|}{}                                                                       & 0.10            & \multicolumn{1}{c|}{1.11} & \multicolumn{1}{c|}{1.09} & \multicolumn{1}{c|}{1.20} & \multicolumn{1}{c|}{1.21} & \multicolumn{1}{c|}{1.06} & \multicolumn{1}{c|}{1.16} & \multicolumn{1}{c|}{1.28} & \multicolumn{1}{c|}{1.27} & \multicolumn{1}{c|}{1.22} & \multicolumn{1}{c|}{1.21} & \multicolumn{1}{c|}{1.19}  & 1.19  & \multicolumn{1}{c|}{72.02}    & 68.84    \\
\multicolumn{1}{c|}{}                                                                       & 0.15            & \multicolumn{1}{c|}{1.09} & \multicolumn{1}{c|}{1.06} & \multicolumn{1}{c|}{1.23} & \multicolumn{1}{c|}{1.13} & \multicolumn{1}{c|}{1.06} & \multicolumn{1}{c|}{1.13} & \multicolumn{1}{c|}{1.18} & \multicolumn{1}{c|}{1.23} & \multicolumn{1}{c|}{1.19} & \multicolumn{1}{c|}{1.18} & \multicolumn{1}{c|}{1.17}  & 1.15  & \multicolumn{1}{c|}{68.28}    & 70.80    \\
\multicolumn{1}{c|}{}                                                                       & 0.20            & 1.09                      & \multicolumn{1}{c|}{1.08} & 1.12                      & \multicolumn{1}{c|}{1.17} & 1.06                      & \multicolumn{1}{c|}{1.07} & 1.24                      & \multicolumn{1}{c|}{1.11} & 1.19                      & \multicolumn{1}{c|}{1.17} & 1.15                       & 1.13  & 71.27                         & 70.28    \\ \midrule \midrule
\multicolumn{1}{c|}{\multirow{4}{*}{WizardLM}}                                              & 0.05            & 0.99                      & \multicolumn{1}{c|}{0.99} & 1.14                      & \multicolumn{1}{c|}{1.14} & 1.21                      & \multicolumn{1}{c|}{1.21} & 1.07                      & \multicolumn{1}{c|}{1.07} & 1.17                      & \multicolumn{1}{c|}{1.17} & 1.11                       & 1.11  & 63.09                         & 63.09    \\
\multicolumn{1}{c|}{}                                                                       & 0.10            & \multicolumn{1}{c|}{1.02} & \multicolumn{1}{c|}{1.06} & \multicolumn{1}{c|}{1.16} & \multicolumn{1}{c|}{1.16} & \multicolumn{1}{c|}{1.15} & \multicolumn{1}{c|}{1.18} & \multicolumn{1}{c|}{1.11} & \multicolumn{1}{c|}{1.06} & \multicolumn{1}{c|}{1.12} & \multicolumn{1}{c|}{1.19} & \multicolumn{1}{c|}{1.11}  & 1.13  & \multicolumn{1}{c|}{63.28}    & 63.52    \\
\multicolumn{1}{c|}{}                                                                       & 0.15            & \multicolumn{1}{c|}{0.97} & \multicolumn{1}{c|}{0.97} & \multicolumn{1}{c|}{1.12} & \multicolumn{1}{c|}{1.16} & \multicolumn{1}{c|}{1.24} & \multicolumn{1}{c|}{1.18} & \multicolumn{1}{c|}{1.10} & \multicolumn{1}{c|}{1.13} & \multicolumn{1}{c|}{1.16} & \multicolumn{1}{c|}{1.19} & \multicolumn{1}{c|}{1.11}  & 1.12  & \multicolumn{1}{c|}{61.00}    & 63.50    \\
\multicolumn{1}{c|}{}                                                                       & 0.20            & 0.96                      & \multicolumn{1}{c|}{0.98} & 1.12                      & \multicolumn{1}{c|}{1.14} & 1.24                      & \multicolumn{1}{c|}{1.23} & 1.15                      & \multicolumn{1}{c|}{1.12} & 1.12                      & \multicolumn{1}{c|}{1.17} & 1.10                       & 1.12  & 62.21                         & 64.30    \\ \midrule \bottomrule
\end{tabular}
\end{table}

\clearpage
\twocolumn
\section*{Acknowledgments}

This research was supported by the Key Program of Jiangsu Science Foundation (BK20243012), the Jiangsu Science Foundation (BG2024036), and the Fundamental Research Funds for the Central Universities (022114380023).

\section*{Contribution Statement}

Jia Zhang and Yao Liu contribute equally as co-first authors. Yi Liu and Lan-Zhe Guo serve as co-corresponding authors.

\bibliographystyle{named}
\bibliography{ijcai25}

\end{document}